\documentclass{article}
\usepackage{stmaryrd}
\usepackage{pdfpages}
\usepackage[english]{babel}
\usepackage{float}
\usepackage[letterpaper,top=2cm,bottom=2cm,left=3cm,right=3cm,marginparwidth=1.75cm]{geometry}

\usepackage{amsmath}
\usepackage{graphicx}
\usepackage[colorlinks=true, allcolors=blue]{hyperref}

\title{StudyFormer : Attention-Based and Dynamic Multi View Classifier for X-ray images}
\author{Lucas Wannenmacher  \qquad   Michael Fitzke \qquad Diane Wilson \qquad Andre Dourson \\ \\ Mars Digital Technologies \\ Antech Imaging Services}

\begin{document}
\maketitle

\begin{abstract}
Chest X-ray images are commonly used in medical diagnosis, and AI models have been developed to assist with the interpretation of these images. However, many of these models rely on information from a single view of the X-ray, while multiple views may be available. In this work, we propose a novel approach for combining information from multiple views to improve the performance of X-ray image classification. Our approach is based on the use of a convolutional neural network to extract feature maps from each view, followed by an attention mechanism implemented using a Vision Transformer. The resulting model is able to perform multi-label classification on 41 labels and outperforms both single-view models and traditional multi-view classification architectures. We demonstrate the effectiveness of our approach through experiments on a dataset of 363,000 X-ray images.

\end{abstract}

\section{Introduction}
Accurate and efficient classification of medical images, such as X-ray images, is crucial for the diagnosis and treatment of diseases. Many AI tools for medical imaging rely on models that process a single view of the X-ray image \cite{_all__2021} \cite{rajpurkar2017chexnet}. However, it is common for multiple views to be taken during a patient's visit to the radiologist. In veterinary medicine there is strong evidence that Multi-View thoracic radiographic studies lead to more sensisitive results for possible structured interstitial pulmonary disease, including metastatic disease (\cite{ober2006comparison}).   In this work, we present a model called Studyformer that is capable of taking a variable number of X-ray images as input and performing multi-label classification.

The architecture of Studyformer combines a convolutional neural network with an attention mechanism implemented using a Vision Transformer \cite{16*16} \cite{earlyConvolution}. This approach allows the model to extract relevant features from each view and effectively combine them to improve classification performance. We demonstrate the effectiveness of Studyformer through experiments on a dataset of 363,000 veterinary X-ray images and show that it outperforms both single-view models and traditional multi-view classification architectures. In addition, Studyformer is able to accept a variable number of views in any position, making it highly adaptable to a variety of uses.

The proposed approach is novel in the field of multi-view classification, using a concatenation method and feature map augmentation in combination with an attention mechanism inspired by the success of Vision Transformers. Studyformer significantly improves classification performance compared to existing models such as MVCNN-based models.

\section{Related work}

\paragraph{Multi-View classification.}
In recent years, the problem of multi-view classification has received significant attention in the machine learning community. A common approach for addressing this problem is to use Multi-View Convolutional Neural Network (MVCNN) architectures \cite{MVCNN}, which first process each view independently using a Convolutional Neural Network (CNN), followed by a view pooling step and final classification using either a Multi-Layer Perceptron or another CNN. There have been several studies that have demonstrated the effectiveness of multi-view classification techniques, particularly in the context of X-ray images \cite{lateral1, lateral2}. Our proposed method, StudyFormer, builds upon these existing approaches by introducing an attention-based and dynamic mechanism for combining the information from different views.

\paragraph{Vision Transformers (ViT).} Transformers (ViTs), introduced in the influential paper "Attention is All You Need" \cite{allYouNeed}, have rapidly gained popularity in the field of Natural Language Processing and have emerged as a promising architecture for a variety of tasks. Transformers were adapted for use in Computer Vision tasks \cite{16*16}, and have subsequently achieved state-of-the-art performance in image classification \cite{distfomer}. In a Vision Transformer (ViT) model, images are decomposed into a sequence of patches which are transformed into tokens and processed by the transformer. Our proposed method, StudyFormer, leverages the effectiveness of ViTs by incorporating them as a key component of our attention-based and dynamic multi-view classifier for X-ray images.

\paragraph{RapidRead} We compare our approach with classification models from \cite{rapidread}, which were trained on large scale medical image data.

\section{Method}
\subsection{Model Architecture}

In this study, we propose a method called StudyFormer for multi-label classification of X-ray images. Our approach uses an adapted Vision Transformer (ViT) that takes as input features extracted from multiple views of the X-ray images. The features are extracted using a Convolutional Neural Network (CNN) that is pretrained on single-view images \cite{rapidread}. The same CNN is applied to each view to extract features, which are then concatenated and augmented to form a square matrix. The square matrix is fed into the ViT to produce the final multi-label classification. Figure \ref{fig:fig1} shows the global architecture of StudyFormer.

\begin{figure}[H]
\centering
\includegraphics[width=14cm]{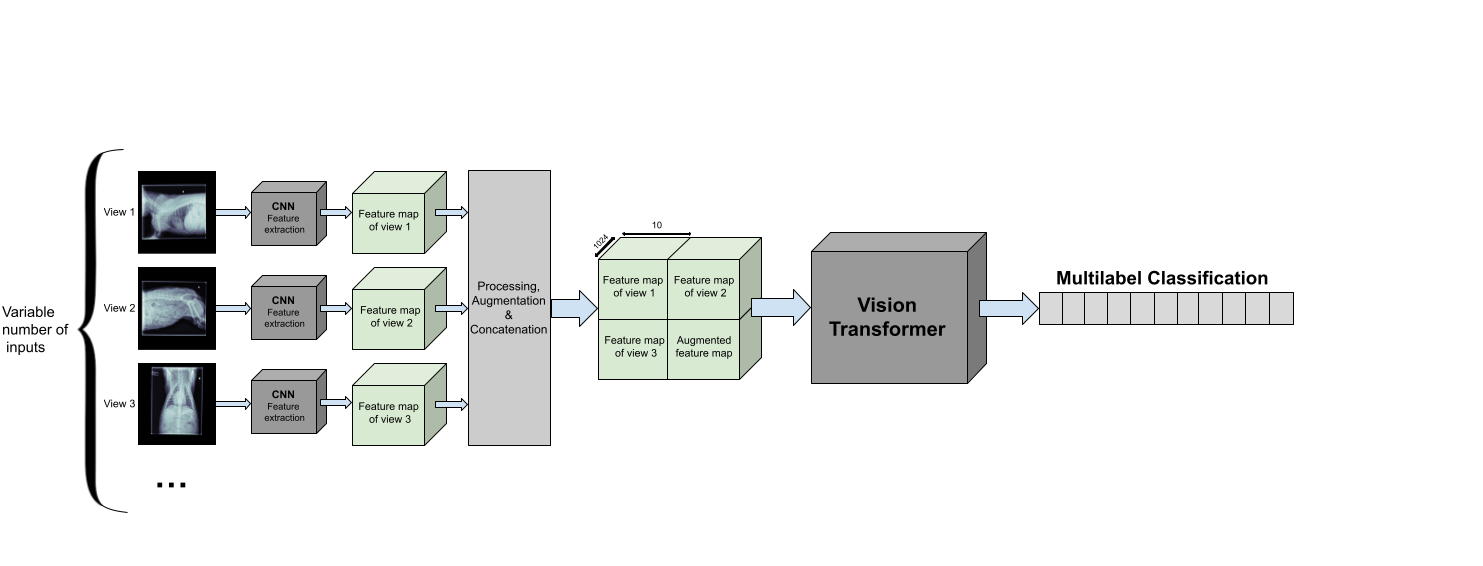}
\caption{StudyFormer architecture}
\label{fig:my_label}
\end{figure}

\subsubsection{Feature Extraction}

The input X-ray views have dimensions of $3\times320\times320$. The CNN used for feature extraction is based on the Densenet121 architecture \cite{DBLP:journals/corr/HuangLW16a} and has pre-trained weights from a model trained for multi-label classification on single views. The output feature maps of the convolutional part have dimensions $10\times10\times1024$.

\subsubsection{Data Augmentation and Square Concatenation}

Data augmentation is used to generate additional feature maps, if needed, to form a square matrix. The feature maps are concatenated to form a square with width $W$, where $W$ is the square root of the total number of feature maps. In our study, the model was trained for $W=2, 3,$ and 4, and can accept up to 16 views ($W=4$).

\subsubsection{ViT}

The input to the ViT is a matrix of size $(W \times 10, W \times 10, 1024)$, which required adaptation of the ViT. We used a patch size of 1, a depth of 6, 16 attention heads, and an MLP dimension of 2048

\subsection{Data}

The data for this work consists of 390850 X-ray images, taken from 98660 veterinary sessions. These images were annotated by radiologists for over 41 diseases in a multi-label fashion, and are feedback from the usage of the RapidRead tool. Additionally, we have a dataset of 800 images with high-quality annotations, where the annotations were performed by 12 radiologists collaborating on each image. A sample of these X-ray images is shown in Figure~\ref{fig:my_label}.

\begin{figure}[H]
\centering
\includegraphics[width=3.8cm]{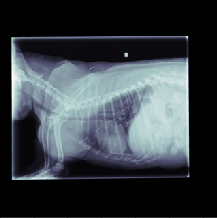}
\includegraphics[width=3.8cm]{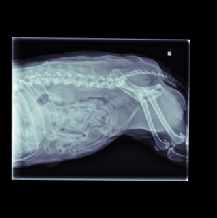}
\includegraphics[width=3.8cm]{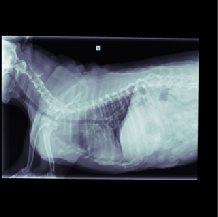}
\includegraphics[width=3.8cm]{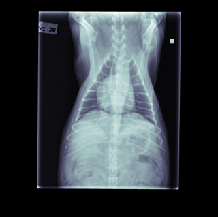}
\caption{Sample X-ray images from the same session}
\label{fig:my_label}
\end{figure}

\subsubsection{Data: Train, Validation and Test Sets}

To ensure unbiased results, we split the X-ray images into three datasets: train, validation, and test, while taking into account the fact that the CNN was previously trained on this dataset. On average, there were $\approx$ 3.96 images per study, which ranges from 1 to more than 10 images sometimes. The total 390850 images corresponded to 98660 studies. The datasets were split as follows:

\begin{itemize}
\item The train set consisted of 363820 images, or 92434 studies, from data prior to April 1, 2022.
\item The validation set consisted of 13515 images, or 3113 studies, from data after April 1, 2022.
\item The test set consisted of 13481 images, or 3113 studies, also from data after April 1, 2022.
\end{itemize}

\subsubsection{Study-Level Labels}

The initial labels were relative to each X-ray, but in this work, we needed labels relative to the studies. Different views of a study often had different labels, as some diseases may be visible in some views and not in others. We made the choice that, for a given label, its value should be the maximum value of that label among all views of the study. Formally, let $L_{i,j,k}$ be the value of the $i^{th}$ label of the $k^{th}$ view of the $j^{th}$ study, and let $LS_{i,j}$ be the study-level label, then:

\begin{align*}
LS_{i,j} &= \max_{k \in \llbracket 1, nb_{views_j}\rrbracket} L_{i,j,k} \
&\forall i \in \llbracket 1, nb_{labels}\rrbracket \
&\forall j \in \llbracket 1, nb_{studies}\rrbracket
\end{align*}

This choice reflects the fact that a patient is considered to have a disease if that disease is detected in at least one view of the study.

\subsubsection{Preprocessing and Transformations}

The preprocessing operations are consistent with those used in the CNN preprocessing pipeline. When switching to a different CNN, the corresponding preprocessing must be applied. The images are first converted to tensors and resized to $\text{3} \times \text{320} \times \text{320}$ ($\text{channels} \times \text{height} \times \text{width}$), followed by normalization. The normalization is performed using the ImageNet mean and standard deviation, which are [0.485, 0.456, 0.406] for the mean and [0.229, 0.224, 0.225] for the standard deviation, respectively.

During training, data augmentation is performed using Horizontal and Vertical Flips, as well as the RandAugment data augmentation method of Pytorch.

\subsubsection{Training}

\subsubsection*{Learning Steps, Computing Power, and Time}

The CNN used was pre-trained for our specific use case, and the first training stage was focused solely on the ViT by freezing the CNN weights. This stage required over 50 epochs and was performed in several phases by saving the weights and optimizer. Subsequently, the entire network was trained, including the CNN, for 10 epochs until the validation loss stopped decreasing.

Most of the training was conducted using a Tesla V100 GPU and took approximately 100 hours.

\begin{figure}[h]
\centering
\includegraphics[width=0.75\linewidth]{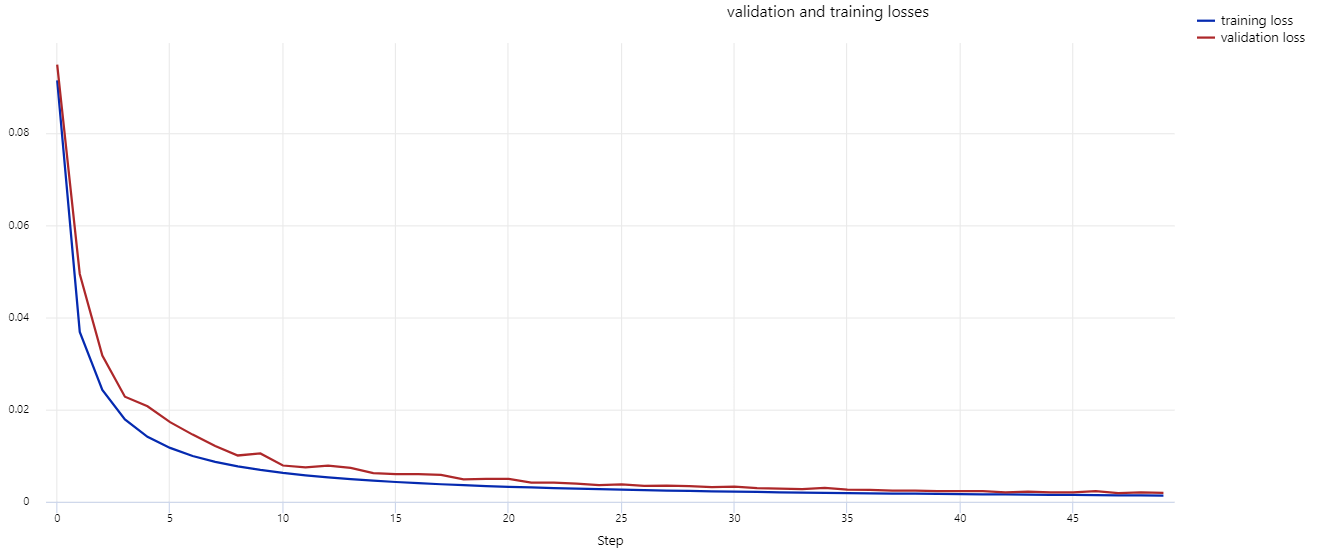}
\caption{Training and validation losses during the ViT-specific training.}
\label{fig:training_validation_losses}
\end{figure}

\subsubsection{Loss function}
Several loss functions, including the BCE (Binary Cross Entropy) and the Focal Loss, were evaluated. The Focal Loss showed promise due to the presence of unbalanced classes, however the results were not better than the BCE. Ultimately, the BCE was chosen as the loss function.

\subsection{Lung Disease Specific Model}

A model was trained specifically for lung-related diseases, with the goal of improving performance on these labels. The model classifies three diseases: Pulmonary Mass, Pulmonary Interstitial Nodole, and Mediastinal Mass Effect.

\subsection{Multi-View-CNN based Architecture}

An alternative architecture, based on the Multi-View-CNN (MVCNN), was also implemented for comparison. In this architecture, the ViT was replaced by a CNN. The results of StudyFormer surpassed the performance of this architecture, as shown in the results section.

\section{Results}

\subsection{Comparison of the metrics between the models}

The performance of the original CNN, the Multi-View-CNN (MVCNN) based architecture, the StudyFormer, and the lung disease specific StudyFormer were compared by evaluating their ROC-AUC scores on different diseases. The ROC-AUC scores of the models are presented in Table 1.

For the single view CNN model, the maximum score over all views was taken as the output of the CNN for each label.

\begin{table}[H]
\centering
\caption{ROC-AUC scores for different models}
\begin{tabular}{ |p{4.8cm}||p{2.2cm}|p{2.2cm}|p{2.2cm}|p{2.2cm}|  }
 \hline
 \multicolumn{5}{|c|}{ROC-AUC scores for different models} \\
 \hline
 Diseases & Single View CNN (Max over views)&MVCNN-based model&StudyFormer &lung-specific StudyFormer \\
 \hline
 Mediastinal Mass Effect &  0.941 & 0.943 & 0.953 & \textbf{0.962}\\
 Pulmonary Mass & 0.929 & 0.928 & 0.941 & \textbf{0.945}\\
 Pulmonary Interstitial - Nodule & 0.905 & 0.908 & \textbf{0.922} & 0.919\\
 Sign(s) of IVDD - Nodule &  0.913 & 0.915 & \textbf{0.940} & /\\
 Gastric Foreign Material (debris) - Nodule & 0.865 & 0.864 & \textbf{0.898} & /\\
 Degenerative Joint Disease - Nodule & 0.838 & 0.839 & \textbf{0.848} & /\\
 Sign(s) of Pleural Effusion - Nodule & 0.935 & 0.935 & \textbf{0.940} & /\\
 Pneumothorax & 0.926 & 0.927 & \textbf{0.959} & /\\
 Pulmonary Vascular &  0.928 & 0.928 & \textbf{0.955} & /\\
 \hline
\end{tabular}
\end{table}

\subsubsection{Comparison of ROC curves}
\begin{figure}[H]
    \centering
    \includegraphics[width=10cm]{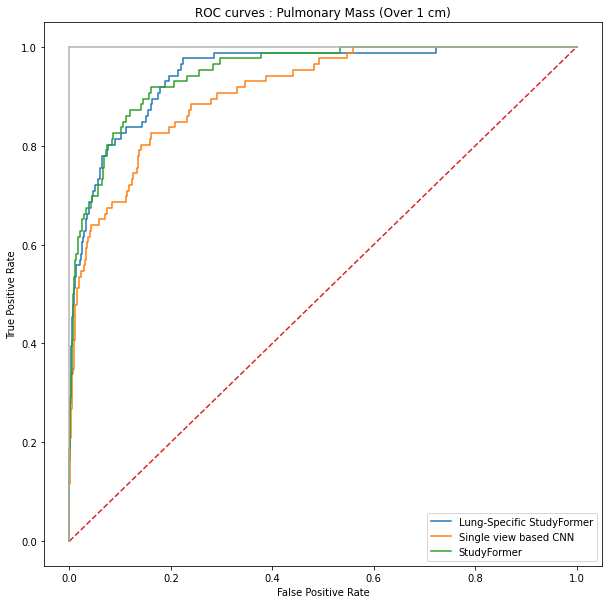}
    \caption{Comparison of ROC curves of the Pulmonary Mass label for different models. The Studyformer and specific-Studyformer curves are above the single view CNN curve.}
\end{figure}

\subsection{Attention map visualisation}

This section presents visualizations of the Vision Transformer (ViT) attention maps. The ViT used is specific to lung diseases, and the attention maps are shown for patients with the positive label 'Pulmonary Mass'. The input to the ViT is a concatenated feature map, and the X-ray images have been mapped and displayed with the same concatenation and transformations applied to the augmented feature maps.

The attention maps show that the ViT focuses on the thorax region where the lungs are located, as expected. The results also demonstrate that the ViT remains focused on the thorax area even with different contexts in the X-ray images. This highlights the robustness of ViTs.

\begin{figure}[H]
    \centering
    \includegraphics[width=7.6cm]{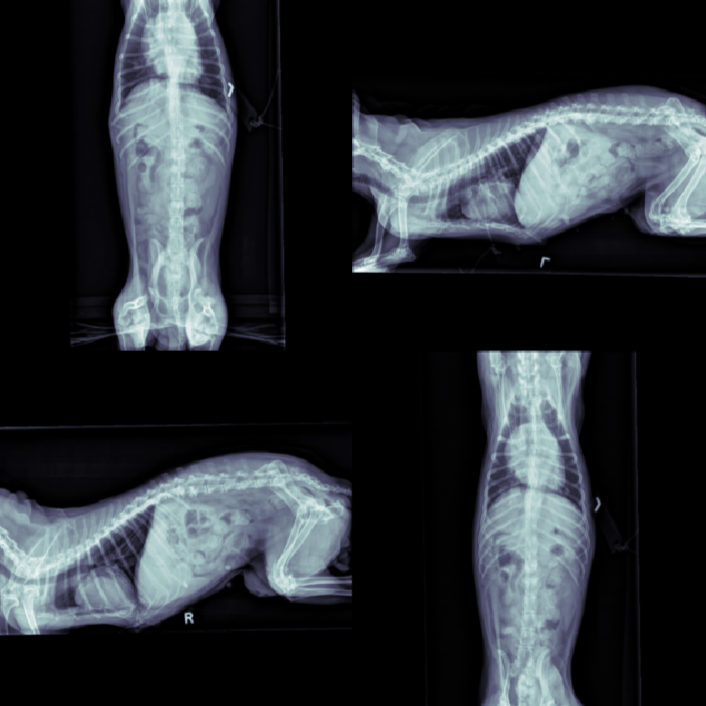}
    \includegraphics[width=7.6cm]{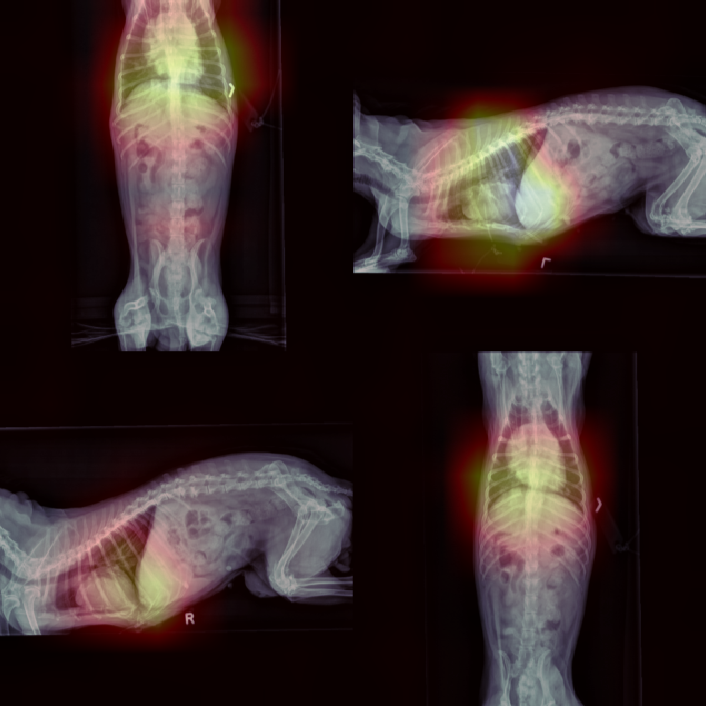}
    \caption{Attention map of the specific-to-lung-diseases model, of a patient with pulmonary mass. We can see here that the attention is focused on the thorax in which the lungs are located, as expected.}
\end{figure}

\begin{figure}[H]
    \centering
    \includegraphics[width=7.6cm]{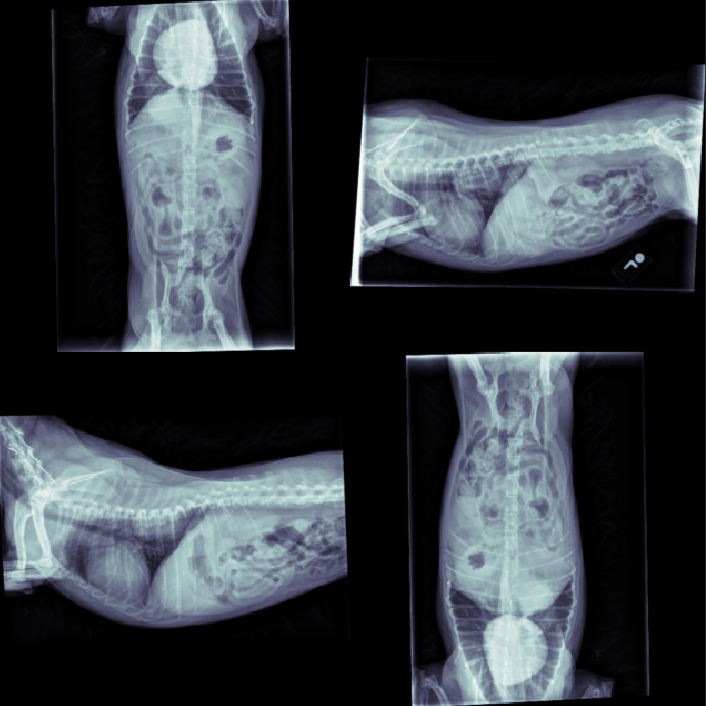}
    \includegraphics[width=7.6cm]{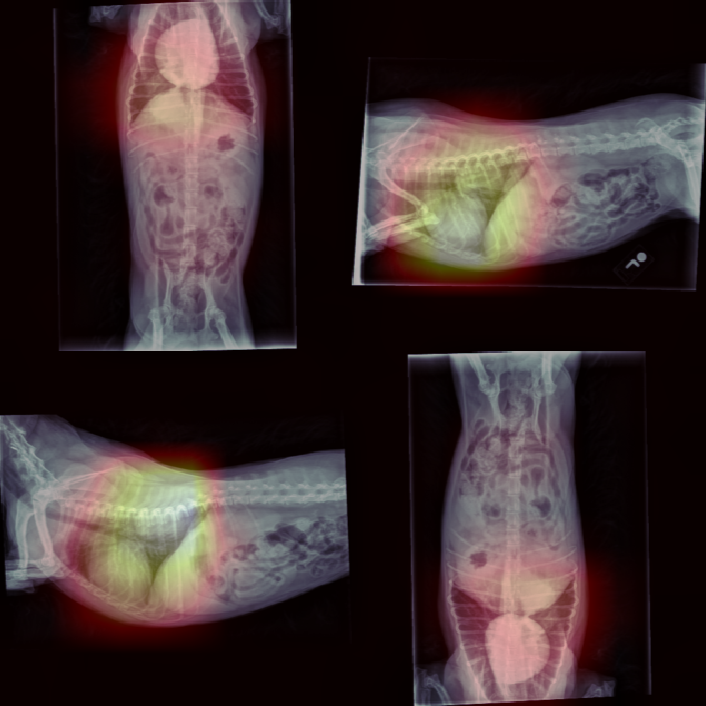}
    \caption{Here too, the focus is on the thorax. Moreover, for the view on the bottom left, the view was created from the 1st image, with some data augmentation: there were initially only 3 views in the study}
\end{figure}

\begin{figure}[H]
    \centering
    \includegraphics[width=7.6cm]{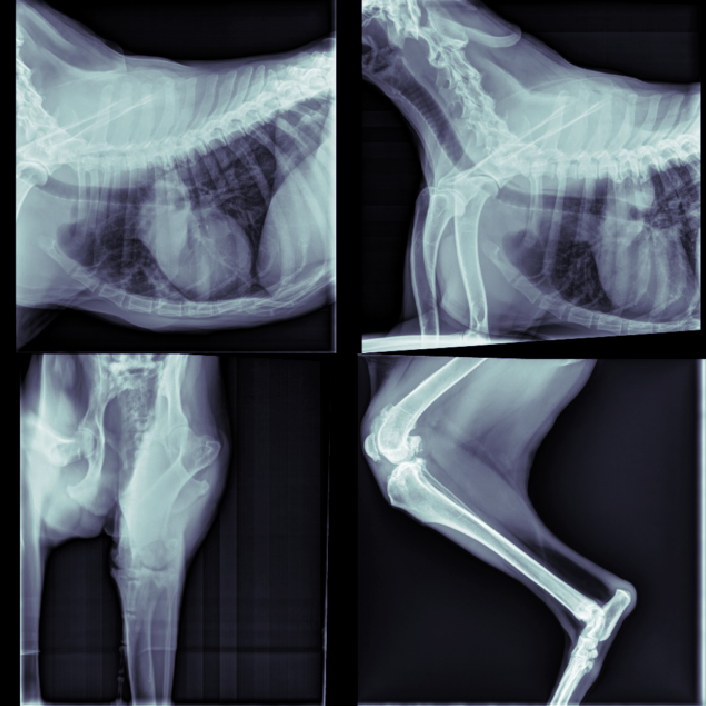}
    \includegraphics[width=7.6cm]{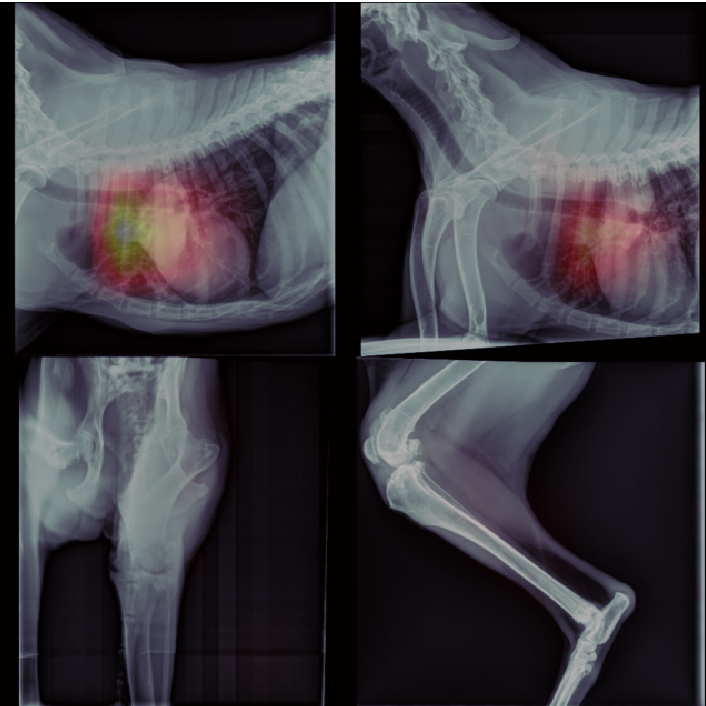}
    \caption{The 3rd and 4th images do not show the lungs area : the ViT does not pay attention to these images, as expected.}
\end{figure}

\begin{figure}[H]
    \centering
    \includegraphics[width=7.6cm]{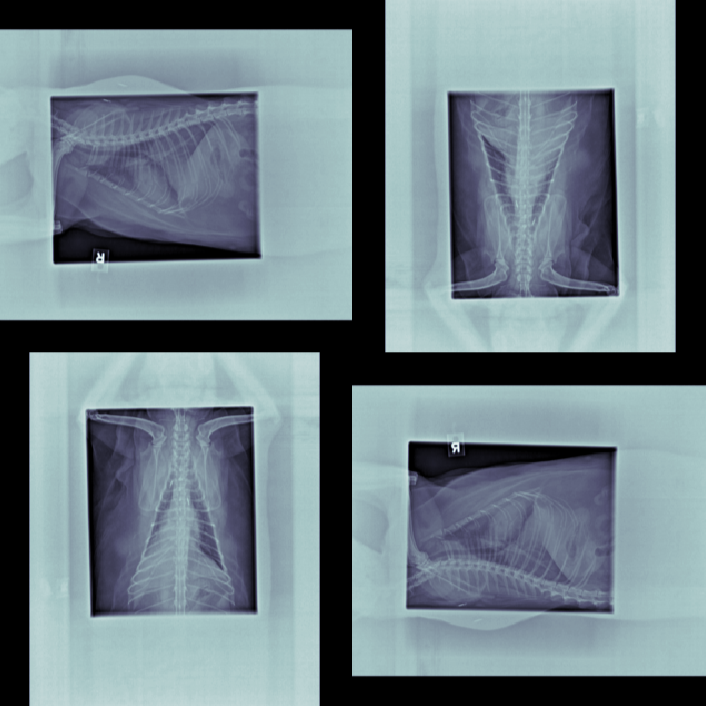}
    \includegraphics[width=7.6cm]{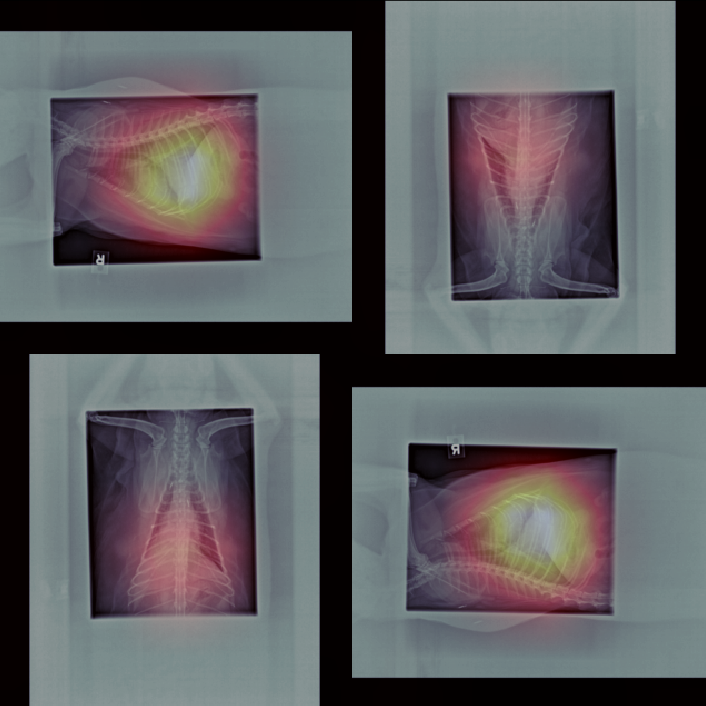}
    \caption{In this study, the context of the images is different. This does not disturb the focus of the ViT, which is still on the thorax area. ViTs are known to be robust.}
\end{figure}

\section{Conclusion}

Our experimental results demonstrate that the Studyformer architecture outperforms single view based architectures in terms of classification performance. The results also show that Studyformer, with its unique architecture that allows a variable number of views to be input in an arbitrary order, performs better than the adapted network in MVCNN under the tested conditions and hyper-parameters.

The visualisation of the Vision Transformer (ViT) attention maps supports the conclusion that the attention is focused on the correct regions in the images and highlights the robustness of the network. This finding is particularly relevant for medical imaging, where the focus on specific regions of the images can be critical for accurate diagnosis.

Furthermore, we observe that using a network specific to a certain group of labels, as in our case with the network specific to lung diseases, may lead to improved performance. This conclusion requires further investigation, but it highlights the potential for specialising networks for specific domains to achieve better results.

In summary, our work contributes to the advancement of multi-view networks and sheds light on the potential of using domain-specific networks for medical image classification.

\bibliographystyle{abbrv}
\bibliography{references}
\end{document}